\documentclass[10pt, a4paper]{article}

\usepackage[final]{lrec2026} 

\usepackage{multirow}
\usepackage[utf8]{inputenc}
\usepackage[T5]{fontenc} 
\usepackage[vietnamese,english]{babel}
\usepackage{amsmath}
\usepackage{amssymb}  
\usepackage{microtype}
\usepackage{url}
\usepackage{booktabs}

\UseMicrotypeSet[protrusion]{basicmath} 
\microtypesetup{protrusion=false}

\title{Tarab: A Multi-Dialect Corpus of Arabic Lyrics and Poetry}

\name{Mo El-Haj} 

\address{College of Engineering and Computer Science, VinUniversity \\
     {\small \texttt{elhaj.m@vinuni.edu.vn}}\\
}

\abstract{We introduce the Tarab Corpus, a large-scale cultural and linguistic resource that brings together Arabic song lyrics and poetry within a unified analytical framework. The corpus comprises 2.56 million verses and more than 13.5 million tokens, making it, to our knowledge, the largest open Arabic corpus of creative text spanning both classical and contemporary production. Tarab is broadly balanced between songs and poems and covers Classical Arabic, Modern Standard Arabic (MSA), and six major regional varieties: Egyptian, Gulf, Levantine, Iraqi, Sudanese, and Maghrebi Arabic. The artists and poets represented in the corpus are associated with 28 modern nation states and multiple historical eras, covering over fourteen centuries of Arabic creative expression from the Pre-Islamic period to the twenty-first century. Each verse is accompanied by structured metadata describing linguistic variety, geographic origin, and historical or cultural context, enabling comparative linguistic, stylistic, and diachronic analysis across genres and time. We describe the data collection, normalisation, and validation pipeline and present baseline analyses for variety identification and genre differentiation. The dataset is publicly available on HuggingFace at \url{https://huggingface.co/datasets/drelhaj/Tarab}.}

\begin{document}

\maketitleabstract

\section{Introduction}
Arabic is characterised by rich linguistic variation across geography, social context, and historical period. Modern Arabic exists in a continuum between Modern Standard Arabic (MSA) and diverse regional dialects, each with distinct phonological, morphological and lexical properties \cite{habash2010introduction}. Dialectal Arabic has received increasing attention in recent years due to its prevalence in real-world communication and the limitations of resources focused only on MSA \cite{zaidan-callisonburch-2014-arabic, el2018arabic, bouamor-etal-2019-madar}. However, most existing Arabic corpora are drawn from news, Wikipedia, or social media, leaving creative forms of language such as song lyrics and poetry significantly underrepresented \cite{mohammad-etal-2016-arabic, obeid2020camel, el2024multilingual}.

Song lyrics and poetry are valuable for Arabic NLP because they encode features that are often absent from standard corpora, including rhyme, metre, emotional expression, repetition, discourse parallelism, and genre-specific conventions. These genres frequently include dialectal intensity, morphological variation, and non-standard orthography \cite{darwish-2014-arabizi, habash2010introduction}, as well as code-switching between Arabic varieties and other languages \cite{habash2014multidialectal,el2024multilingual}. Poetry also captures Classical Arabic forms across historical eras, offering opportunities for diachronic linguistic analysis \cite{al2020meter,qarah2024arapoembert}. Despite this linguistic richness, there is currently no large-scale, publicly available corpus that unifies both Arabic song lyrics and poetry in a way that supports comparative analysis across dialects, genres, and historical periods.

This paper introduces the Tarab Corpus, a large-scale resource of Arabic creative language encompassing both song lyrics and poetry across modern and historical contexts\footnote{\url{https://huggingface.co/datasets/drelhaj/Tarab}}. Tarab, often translated as musical ecstasy or aesthetic rapture, refers to a culturally grounded affective state of deep emotional engagement experienced in Arabic musical and poetic traditions. The corpus comprises 2,557,311 verses and 13,509,336 tokens, with each verse annotated for linguistic variety, geographic origin, and historical or cultural context. Tarab spans texts from contemporary popular music and modern poetry to classical literary traditions associated with major historical eras, capturing Arabic language use across time, region, and genre. In contrast to existing resources that are typically restricted to a single variety or domain \cite{zaidan-callisonburch-2014-arabic, bouamor-etal-2019-madar}, Tarab enables cultural, computational, and sociolinguistic research at a scale and level of diversity not previously available.

\section{Related Work}

Arabic language resources have expanded in recent years, yet most available corpora focus on news and encyclopaedic text \cite{el2013kalimat, antoun-etal-2020-arabert}. Major efforts such as the Arabic Gigaword Corpus \cite{parker2011arabic} and the OSIAN web corpus \cite{zerrouki-bouamor-2019-osian} support large-scale modelling of Modern Standard Arabic (MSA) but do not address dialectal or creative linguistic forms. With the rise of interest in Arabic dialect processing, several dialectal corpora have been introduced, including the Arabic Online Commentary dataset \cite{zaidan-callisonburch-2011-arabic}, the MADAR corpus of parallel sentences across Arabic cities \cite{bouamor-etal-2019-madar}, and country-level social media corpora \cite{alhazmi2024code, mubarak-etal-2021-arabic}. These resources enabled progress in dialect identification, but they are limited to prose and do not represent verse or musical language.

Work on Arabic poetry and cultural text remains relatively scarce in NLP. Projects such as OpenITI \cite{openiti2019} and Shamela \cite{otakar2014} have made important progress in digitising classical Arabic texts, and the AlKhalil morphological analyser for Classical Arabic \cite{boudlal2010alkhalil,boudchiche2017alkhalil} enables heritage text analysis. However, these collections focus primarily on prose rather than poetry or song. Arabic poetry has been studied computationally in the context of metre classification \cite{al2020meter, mutawa2025determining}, but available datasets are small in scale and constrained to classical forms. There remains a gap in large unified poetic corpora that also include modern verse and dialectal variation.

Song lyrics represent another creative domain that reflects informal language and dialectal richness, but they are significantly underrepresented in Arabic NLP. Lyrics exhibit features such as rhyme, repetition and colloquial morphology, making them useful for studying linguistic variation and stylistic modelling. \citet{elhaj2020habibi} introduced the Habibi Lyrics Corpus , one of the first Arabic lyrics resources covering multiple dialects. That work demonstrated the value of lyrics for dialect identification, but it was limited to musical content and did not include poetry or historical linguistic dimensions.

The Tarab Corpus builds on this line of research by extending the scope of creative Arabic resources beyond lyrics to also include poetry. Unlike previous datasets, Tarab integrates both modern and classical text, linking verse-level entries to dialect, origin and historical metadata. This makes it possible to study variation across genre, geography and historical period within a single framework. To our knowledge, this is the first Arabic resource to unify lyrics and poetry at scale for linguistic, cultural and computational analysis.

\section{Corpus Creation and Design}
\label{sec:corpus}

The Tarab Corpus is a large-scale resource of Arabic creative expression that brings together song lyrics and poetry within a single, unified framework. Rather than treating these genres as separate cultural artefacts, Tarab adopts the verse as its basic unit of analysis, enabling systematic comparison across genre, linguistic variety, geography, and historical period. This design supports analyses that span performance, literature, and orality, which are difficult to conduct using existing Arabic resources that focus primarily on prose or single varieties.

Tarab captures both contemporary and heritage forms of Arabic creativity. It combines a broad spectrum of song lyrics drawn from popular, folk, and religious repertoires with a substantial body of Arabic poetry ranging from early literary traditions to modern poetic practice. In total, the corpus comprises 2,557,311 verses and more than 13.5 million tokens, representing 89,166 distinct works produced by 2,598 unique creators (2,060 singers and 538 poets) associated with 28 modern countries and major historical eras, from the Pre-Islamic period through successive Islamic dynasties to the present. Linguistic coverage spans Classical Arabic, Modern Standard Arabic (MSA), and six major regional dialect groups, supporting research that connects Arabic literary heritage with contemporary popular culture.

Tarab is constructed from three main streams. First, the poetry component builds on an openly available Arabic poetry collection released on Kaggle\footnote{\url{https://www.kaggle.com/datasets/ahmedabelal/arabic-poetry}}. Second, the lyrics component includes material from the Habibi corpus \cite{elhaj2020habibi}. Third, we extend coverage by crawling additional publicly accessible web pages containing lyric text. Crawling was restricted to sites that permit automated access, operationalised by checking that the site's \texttt{robots.txt} does not disallow retrieval of the relevant paths. The final dataset is represented uniformly at verse level, with all sources normalised into the same schema described in Section~\ref{subsec:schema}.

\subsection{Creative scope}
\label{subsec:scope}

Tarab draws from two primary creative domains: song lyrics and poetry. The lyrics component spans a wide range of stylistic and cultural contexts rather than fixed, explicitly annotated genres. These include mainstream popular songs, religious (\emph{dini}) material, hip-hop and rap, and songs associated with particular musical styles or performance traditions such as Khaleeji and Maghrebi. These stylistic categories are not treated as mutually exclusive labels tied to artist nationality or dialect. For instance, a song may be performed by an artist from Tunisia, contain Maghrebi dialectal features, and yet follow a Khaleeji musical style. Such distinctions are preserved through separate metadata fields and auxiliary resources rather than collapsed into a single genre label.

The poetry component includes both contemporary poetry and heritage poetry. Contemporary poems are associated with modern national origins, such as Iraq, the United Arab Emirates, or Palestine, while heritage poetry is linked to major historical periods including the Abbasid, Ayyubid, Andalusian, Mamluk, and Ottoman eras. This dual representation enables the study of poetic language across both modern sociocultural contexts and long-term historical trajectories. Together, the two domains provide a continuous view of Arabic creative language across performance traditions, registers, and time, while allowing dialect, style, and origin to be analysed independently.

\subsection{Verse-level representation and schema}
\label{subsec:schema}

All content in Tarab is represented using a unified verse-level schema. Each verse occupies a single row and is linked to its parent work through stable identifiers, allowing both fine-grained linguistic analysis and reconstruction of full songs or poems when needed. The schema includes the following fields: \texttt{art\_id}, \texttt{artist\_id}, \texttt{artist\_name}, \texttt{art\_title}, \texttt{writer}, \texttt{composer}, \texttt{verse\_order}, \texttt{verse\_lyrics}, \texttt{origin} (modern country or historical era), \texttt{dialect}, and \texttt{type} (song or poem). This representation supports longitudinal analysis, cross-genre comparison, and reproducible experimentation across linguistic varieties and historical periods.

\subsection{Pre-processing}
\label{subsec:preproc}

All text in Tarab is stored in UTF-8 and undergoes minimal pre-processing in order to preserve dialectal, orthographic, and stylistic variation. Orthographic features that carry linguistic or regional signal, such as Egyptian \emph{alef maqsura} usage, Gulf vowel elongation, and Maghrebi conventions, are intentionally retained. Verse segmentation follows the line structure of the source material, and the \texttt{verse\_order} field preserves intra-song and intra-poem sequencing. No stemming, lemmatisation, or stopword removal is applied, avoiding the loss of information relevant to linguistic, stylistic, and cultural analysis. 

To ensure internal consistency and prevent duplication, works are identified and validated using a composite key defined over \texttt{(art\_id, artist\_id, verse\_order)}. This allows repeated verses, alternative textual witnesses, and variant performances to be handled systematically while preserving a clear notion of what constitutes a distinct creative work.

\subsection{Corpus composition and growth}
\label{subsec:composition}

Table~\ref{tab:corpus_overview} summarises the composition of the Tarab corpus by genre. While poetry accounts for a larger share of works and verses, the two genres differ in average verse length, reflecting stylistic differences between poetic and musical forms.

\begin{figure}[h]
\centering
\includegraphics[width=0.45\textwidth]{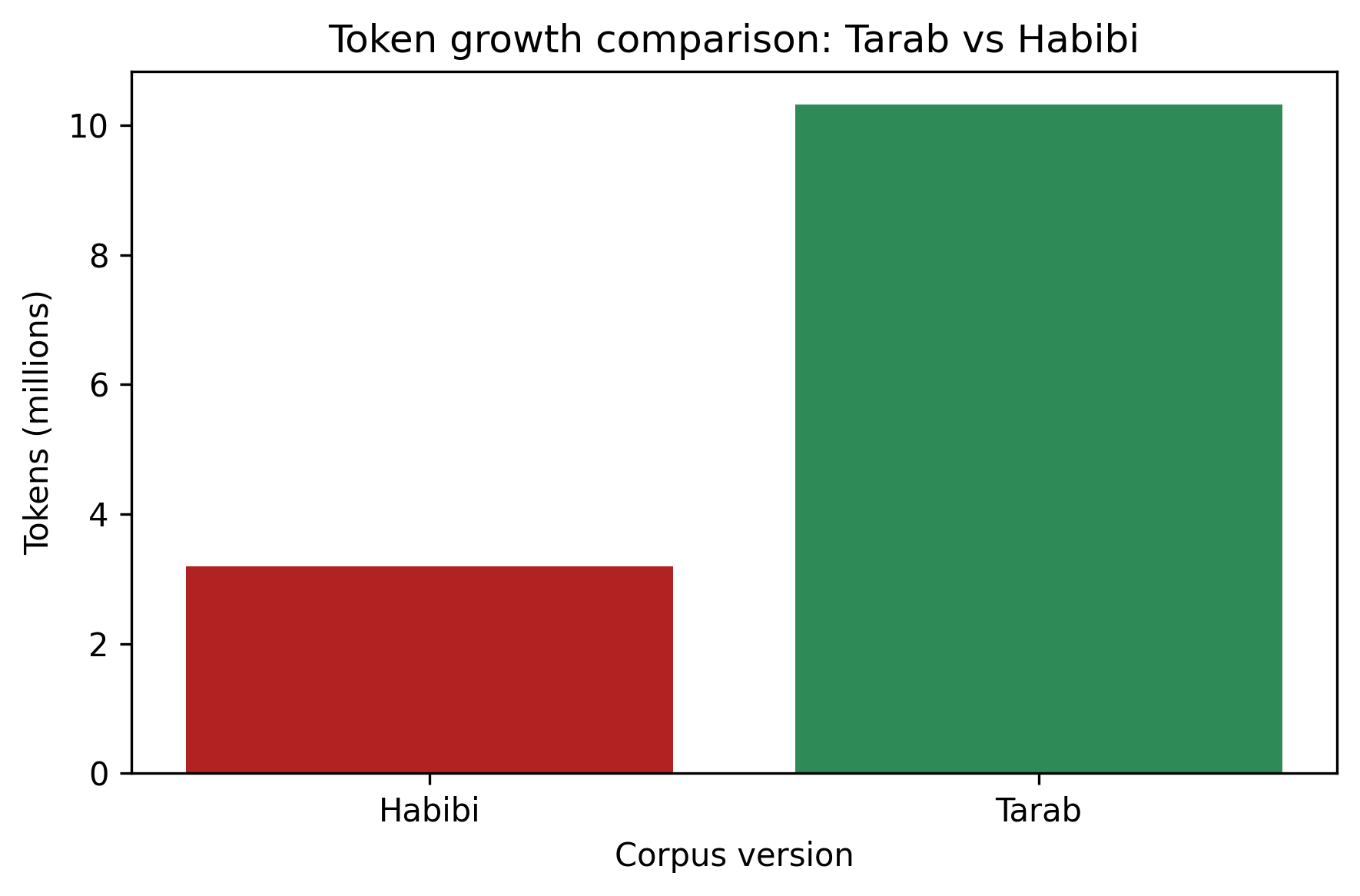}
\caption{Corpus growth in scale compared to earlier Arabic lyrics resources.}
\label{fig:growth}
\end{figure}

Tarab represents a substantial expansion over earlier Arabic lyrics resources. Compared to the Habibi corpus \cite{elhaj2020habibi}, which contains 527,896 lyric verses, Tarab increases the total number of verses by a factor of 4.8, incorporating an additional 1,387,283 verses of poetry alongside 642,221 further lyric verses. This expansion broadens the scope from purely modern musical texts to a unified collection spanning contemporary songwriting and classical Arabic poetics. The \texttt{corpus\_version} field indicates whether a song was originally present in the Habibi corpus, supporting controlled analyses of diachronic and genre variation. Habibi corpus did not include poetry. Figure~\ref{fig:growth} illustrates the relative contribution of each corpus version.

\begin{table}[h]
\centering
\small
\setlength{\tabcolsep}{6pt}
\resizebox{\columnwidth}{!}{%
\begin{tabular}{lrrrr}
\toprule
\textbf{Subset} & \textbf{Works} & \textbf{Verses} & \textbf{Tokens} & \textbf{Avg tokens/verse} \\
\midrule
Songs & 34,239 & 1,170,028 & 6,989,019 & 4.9 \\
Poems & 54,927 & 1,387,283 & 6,520,317 & 5.6 \\
\midrule
\textbf{Total} & \textbf{89,166} & \textbf{2,557,311} & \textbf{13,509,336} & 5.3 \\
\bottomrule
\end{tabular}
}
\caption{Composition of the Tarab corpus by genre, showing the number of works, verses, tokens, and average verse length.}
\label{tab:corpus_overview}
\end{table}

\section{Linguistic, Geographic, and Structural Coverage}
\label{sec:coverage}

This section describes the linguistic, geographic, and structural properties of the Tarab corpus, followed by a detailed analysis of its lexical and stylistic characteristics. Together, these perspectives provide a comprehensive account of how Arabic creative language is distributed, structured, and realised across dialects, genres, and historical contexts.

\subsection{Linguistic and dialectal coverage}
\label{subsec:dialects}

At the linguistic level, Tarab spans Classical Arabic, Modern Standard Arabic (MSA), and six major regional dialect groups. Table~\ref{tab:dialects} summarises the distribution of verses by dialect, together with vocabulary size and average verse length. Classical Arabic and MSA together account for a substantial proportion of the corpus, reflecting the prominence of poetry and formal literary production. In contrast, song lyrics contribute extensive coverage of spoken regional varieties, including Egyptian, Gulf, Levantine, Iraqi, Sudanese, and Maghrebi Arabic, ensuring that contemporary vernacular usage is well represented.

\begin{table}[h]
\centering
\resizebox{\columnwidth}{!}{%
\small
\begin{tabular}{lrrrr}
\hline
Dialect & Verses & Vocab size & Avg tokens/verse & \% of corpus \\
\hline
Classical  & 937,473 & 1,044,325 & 4.7 & 36.7 \\
MSA        & 449,810 & 577,073   & 4.6 & 17.6 \\
Egyptian   & 308,714 & 120,507   & 6.3 & 12.1 \\
Gulf       & 308,249 & 133,599   & 6.1 & 12.1 \\
Levantine  & 250,276 & 119,455   & 5.9 & 9.8 \\
Iraqi      & 156,153 & 73,531    & 5.5 & 6.1 \\
Sudanese   & 89,226  & 58,092    & 5.7 & 3.5 \\
Maghrebi   & 57,410  & 33,762    & 6.0 & 2.2 \\
\hline
\end{tabular}
}
\caption{Vocabulary size and average verse length by dialect in the Tarab corpus.}
\label{tab:dialects}
\end{table}

\subsection{Geographic and historical provenance}
\label{subsec:origins}

\begin{figure}[h]
\centering
\includegraphics[width=0.45\textwidth]{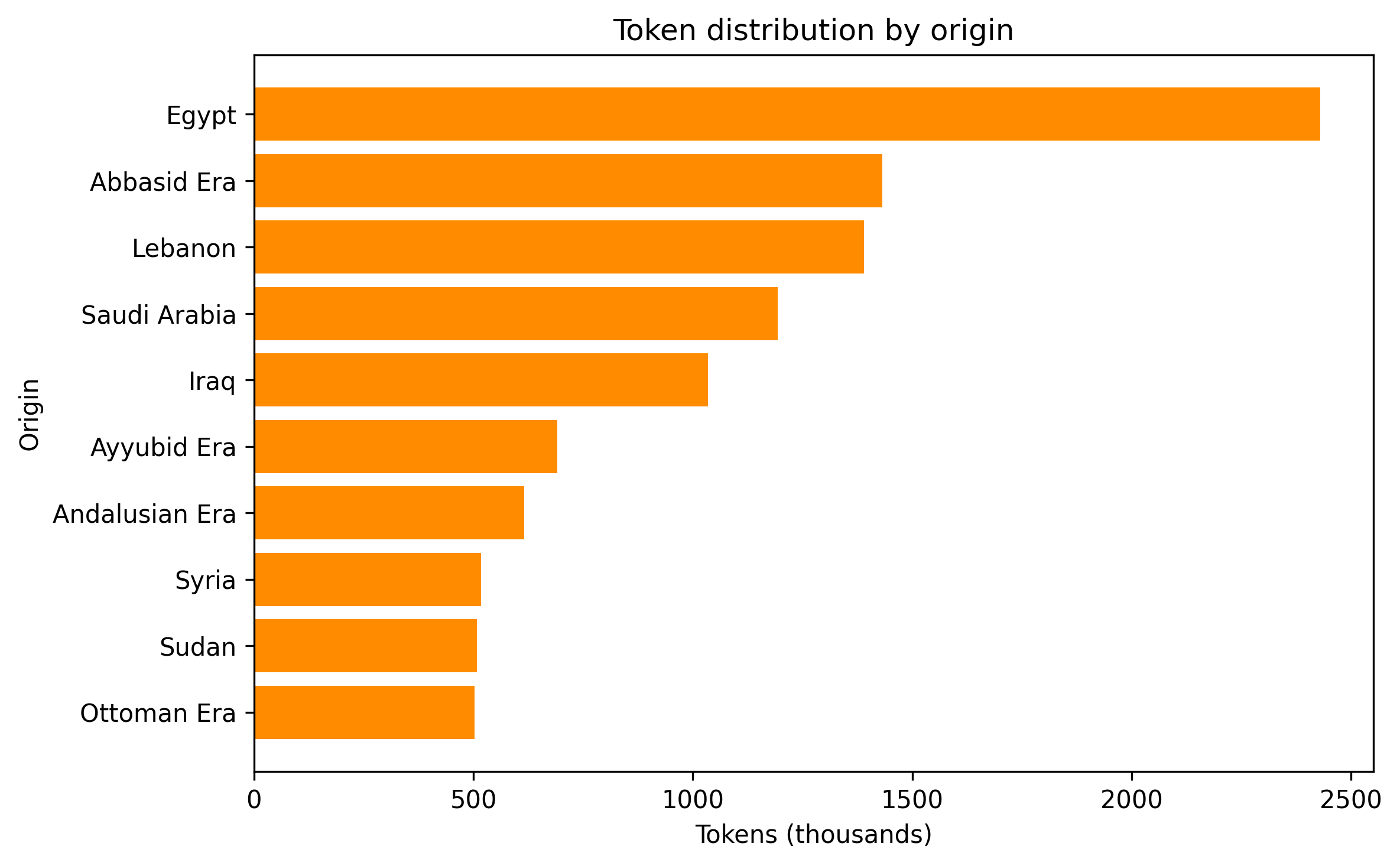}
\caption{Top origins by verse count, including modern countries and historical eras.}
\label{fig:origins}
\end{figure}

Tarab incorporates material associated with both modern nation states and major historical eras, spanning over fourteen centuries of Arabic creative text, from pre-610 CE poetry to contemporary songs and modern poetic production in the twenty-first century. Figure~\ref{fig:origins} shows the most prominent origins by verse count. Modern countries such as Egypt, Lebanon, and Saudi Arabia contribute a large share of song lyrics, while historical periods including the Abbasid, Andalusian, and Mamluk eras account for a substantial proportion of the poetic material. This explicit separation between geographic origin and historical era enables analysis across time and space without conflating linguistic variety with chronology.

Table~\ref{tab:origins} presents the full distribution of works, tokens, and verses across modern countries and historical eras.

\begin{table}[h]
\centering
\small
\resizebox{\columnwidth}{!}{%
\begin{tabular}{lrrr}
\toprule
\textbf{Origin} & \textbf{Works} & \textbf{Tokens} & \textbf{Verses} \\
\midrule
Egypt & 11,182 & 2,429,198 & 414,914 \\
Abbasid Era & 13,456 & 1,431,613 & 303,378 \\
Lebanon & 7,390 & 1,390,369 & 253,143 \\
Saudi Arabia & 6,575 & 1,193,549 & 197,384 \\
Iraq & 4,913 & 1,034,427 & 195,165 \\
Ayyubid Era & 5,018 & 690,972 & 143,768 \\
Andalusian Era & 4,410 & 616,022 & 130,040 \\
Ottoman Era & 3,937 & 502,892 & 108,743 \\
Mamluk Era & 6,095 & 490,866 & 102,999 \\
Syria & 2,820 & 517,833 & 99,693 \\
Sudan & 2,683 & 507,783 & 89,829 \\
Kuwait & 1,962 & 361,052 & 61,867 \\
Palestine & 1,429 & 271,712 & 56,448 \\
United Arab Emirates & 1,719 & 310,004 & 54,462 \\
Islamic Era & 2,351 & 264,482 & 54,081 \\
Morocco & 1,259 & 235,739 & 41,298 \\
Era of the Mukhadramun & 2,167 & 192,953 & 40,692 \\
Pre-Islamic Era & 1,989 & 175,622 & 36,826 \\
Tunisia & 1,072 & 168,709 & 31,671 \\
Yemen & 1,360 & 153,797 & 30,535 \\
Algeria & 807 & 129,197 & 25,157 \\
Umayyad Era & 2,360 & 124,200 & 24,817 \\
Jordan & 775 & 125,656 & 23,574 \\
Oman & 872 & 95,100 & 19,872 \\
Bahrain & 207 & 35,515 & 5,863 \\
Qatar & 199 & 33,696 & 5,723 \\
Libya & 133 & 18,292 & 3,775 \\
Mauritania & 27 & 8,086 & 1,594 \\
\midrule
\textbf{Total} & \textbf{89,166} & \textbf{13,509,336} & \textbf{2,557,311} \\
\bottomrule
\end{tabular}
}
\caption{Distribution of works, tokens, and verses across modern countries and historical eras.}
\label{tab:origins}
\end{table}

\subsection{Structural properties of verses}
\label{subsec:structure}

\begin{figure}[h]
\centering
\includegraphics[width=0.45\textwidth]{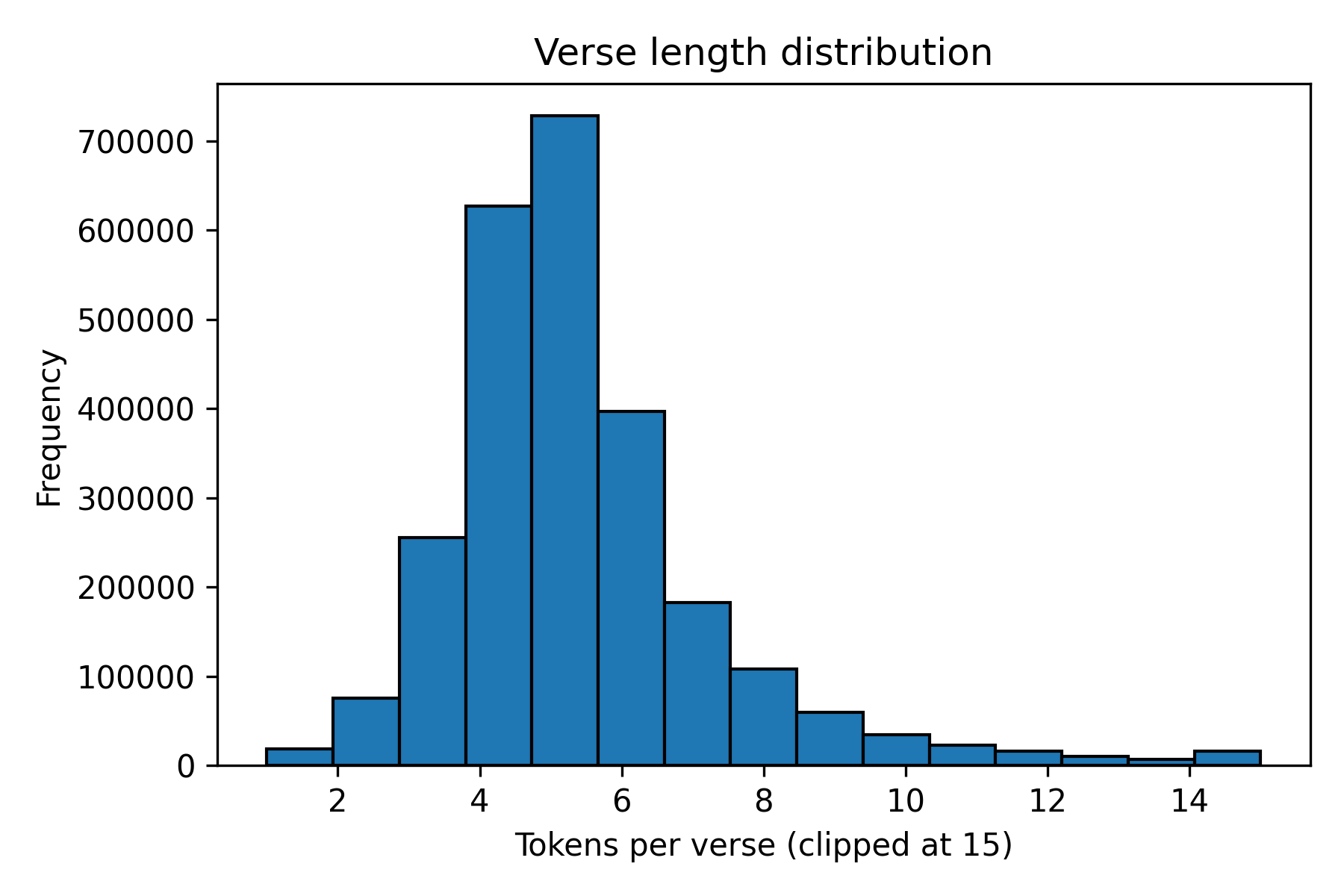}
\caption{Distribution of verse lengths across the Tarab corpus.}
\label{fig:verse_length}
\end{figure}

At a structural level, verses in Tarab are typically short. Figure~\ref{fig:verse_length} shows the distribution of tokens per verse, with most verses falling between three and eight tokens.

\subsection{Dialectal lexical variation}
\label{subsec:lexical_variation}

Given this distributional profile, Tarab exhibits linguistic behaviour that differs markedly from newswire and social media corpora commonly used in Arabic NLP. Lexical choice and tokenisation patterns are shaped by creative constraints, including metre, repetition, and performance, rather than sentence-based prose structure.

Dialectal variation is particularly visible in vocabulary composition. Classical Arabic and MSA display the largest vocabularies, consistent with the lexical richness and stylistic range of poetic language. Regional dialects, while smaller in vocabulary size, exhibit strong lexical distinctiveness and longer average verse lengths, especially in song lyrics.

Beyond aggregate statistics, Tarab exhibits clear and systematic dialectal differentiation that reflects regionally grounded usage across the corpus. This diversity is evident in the high-frequency lexical items summarised in Table~\ref{tab:top_tokens_examples}, which highlight recurrent dialect-specific forms rather than shared pan-Arabic vocabulary. Across regional varieties, these items include characteristic discourse particles, address forms, and affective expressions that are widely attested in spoken interaction and creative language.

For instance, Maghrebi varieties show frequent use of forms such as \textit{ElA\$} (why), \textit{bgyt} (I want), and \textit{mAzAl} (still), which are strongly associated with Maghrebi Arabic. Similarly, Gulf Arabic is characterised by vocative and affective expressions such as \textit{wynk} (where are you) and \textit{yA bEdy} (my beloved), while Egyptian and Levantine varieties exhibit colloquial particles and pronominal forms typical of everyday speech. Together, the patterns illustrated in Table~\ref{tab:top_tokens_examples} demonstrate that Tarab captures robust dialectal influence across regions, strengthening the corpus's linguistic diversity and its suitability for research on dialect modelling and regional stylistics.

\begin{table}[h]
\centering
\small
\resizebox{\columnwidth}{!}{%
\begin{tabular}{lll}
\hline
\textbf{Dialect} & \textbf{Common lexical items (Buckwalter)} & \textbf{Gloss (English)} \\
\hline
Classical  & qAl, yA, <in~, ElY, mA & said, O/oh, indeed, on/upon, what \\
MSA        & Al*y, h*h, kAn, <ilY, En & which/that, this, was, to, about \\
Egyptian   & lyh, bs, qlby, dh, <nt & why, just, my heart, this, you \\
Gulf       & wynk, yA bEdy, >Hb~k, AlglA & where are you, my beloved, I love you, dear \\
Levantine  & \$w, Hbyby, hyk, ls~, qlby & what, my love, like this, still, my heart \\
Iraqi      & \$lwn, >Any, rwHy, <nt, wyn & how, I, my soul, you, where \\
Sudanese   & xlAS, mAlk, HbAbk, wyn, slAm & enough/ok, what's wrong, welcome, where, peace \\
Maghrebi   & bgyt, mAzAl, ElA\$, Hby, qlby & I want, still, why, my love, my heart \\
\hline
\end{tabular}
}
\caption{Examples of frequent lexical items by dialect with English glosses using Buckwalter transliteration.}
\label{tab:top_tokens_examples}
\end{table}


\subsection{Code-switching and multilingual influence}
\label{subsec:codeswitch}

Tarab contains natural but unevenly distributed instances of code-switching, overwhelmingly concentrated in song lyrics. Code-switching occurs in approximately 0.6\% of song verses and is virtually absent in poetry. At the artwork level, around 2.3\% of songs contain at least one instance of code-switching, compared to fewer than 0.1\% of poems. Latin-script tokens account for about 0.44\% of all song tokens and are negligible in poetry.

The code-switched material consists primarily of French and English lexical items, particularly in Maghrebi and Lebanese lyrics, including \emph{mon amour}, \emph{baby}, \emph{merci}, and \emph{fiesta}. These patterns align with contemporary sociolinguistic practice in popular music and highlight Tarab's value for studying multilingualism and language contact in Arabic creative contexts.

\subsection{Word-level lexical structure}
\label{subsec:embeddings}

Beyond aggregate statistics, Tarab enables fine-grained analysis of how lexical items associated with different varieties and genres are organised in distributional space. To explore this, we conduct a word-level analysis using FastText embeddings trained on the Tarab corpus. Focusing on word types rather than verses or documents allows us to examine lexical relationships directly, without conditioning on higher-level structural or stylistic units. This is particularly relevant for Arabic, where variation across dialects and registers is often realised at the lexical and morphological level.

FastText \cite{bojanowski2017enriching} is well suited to this setting, as its subword modelling captures morphological variation and orthographic regularities characteristic of both standard and non-standard varieties of Arabic. We retain the full vocabulary when training and analysing the embeddings, allowing frequent and infrequent items alike to contribute to the structure of the space. The resulting word embeddings are projected into two dimensions using t-SNE to support qualitative inspection of how lexical items associated with different varieties and genres are distributed within a shared embedding space.

\begin{figure}[h]
\centering
\includegraphics[width=\linewidth]{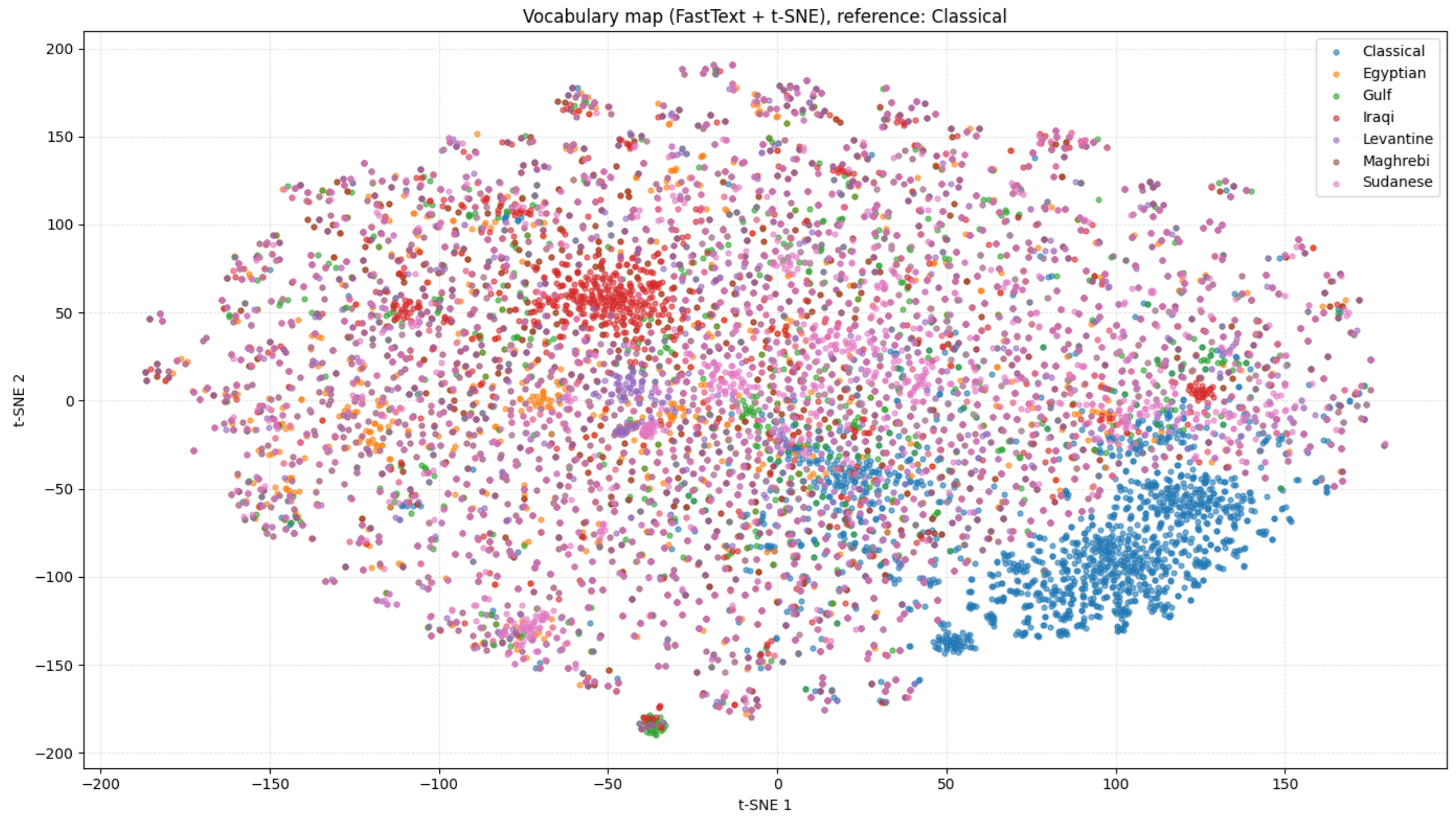}
\caption{Word-level vocabulary map with Classical Arabic as the reference variety.}
\label{fig:classic-tsne}
\end{figure}

Figure~\ref{fig:classic-tsne} visualises the lexical space with Classical Arabic as the reference variety. Classical Arabic forms a compact and largely isolated region, with minimal overlap with dialectal vocabularies. This pattern is consistent with the specialised and genre-bound use of Classical Arabic in Tarab, where its vocabulary tends to occur in constrained poetic and rhetorical contexts that are rarely shared with colloquial varieties.

\begin{figure}[h]
\centering
\includegraphics[width=\linewidth]{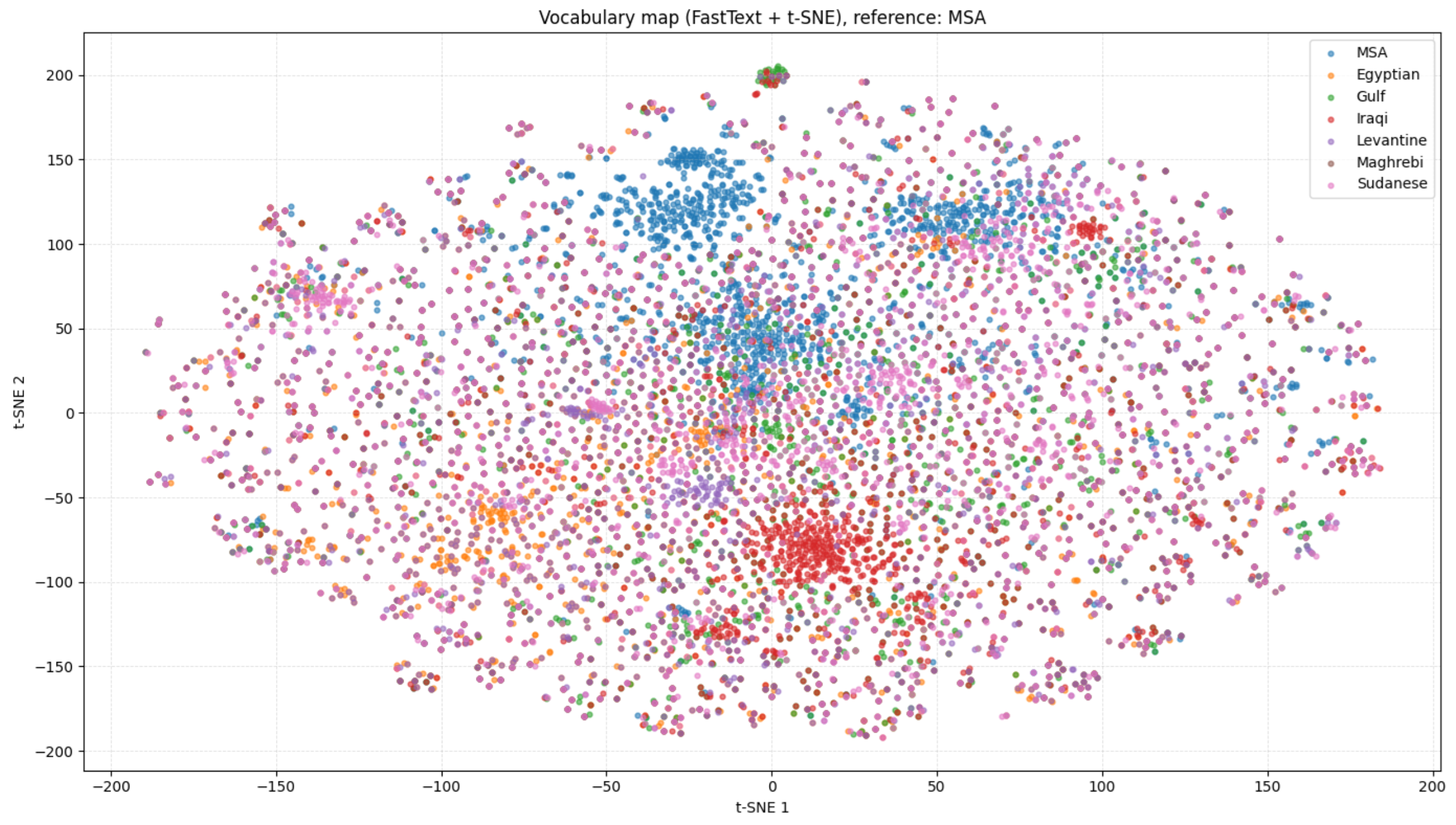}
\caption{Word-level vocabulary map with MSA as the reference variety.}
\label{fig:msa-tsne}
\end{figure}

In contrast, Figure~\ref{fig:msa-tsne} shows Modern Standard Arabic (MSA) occupying a denser and more permeable core of the lexical space. While MSA vocabulary remains internally cohesive, dialectal word forms are distributed around and partially interleaved with it, suggesting substantial lexical sharing and contextual proximity. This organisation aligns with the role of MSA in Tarab as a central written and semi-formal register that co-exists with regional dialects, particularly in song lyrics.

It is important to note that this analysis does not explicitly distinguish between poetic texts and song lyrics. While Classical Arabic in Tarab is predominantly realised in poetry, and MSA material, though often poetic in form, is frequently performed in songs, these genre differences are not encoded in the embedding space and are therefore conflated in the visualisation.

\begin{figure}[h]
\centering
\includegraphics[width=\linewidth]{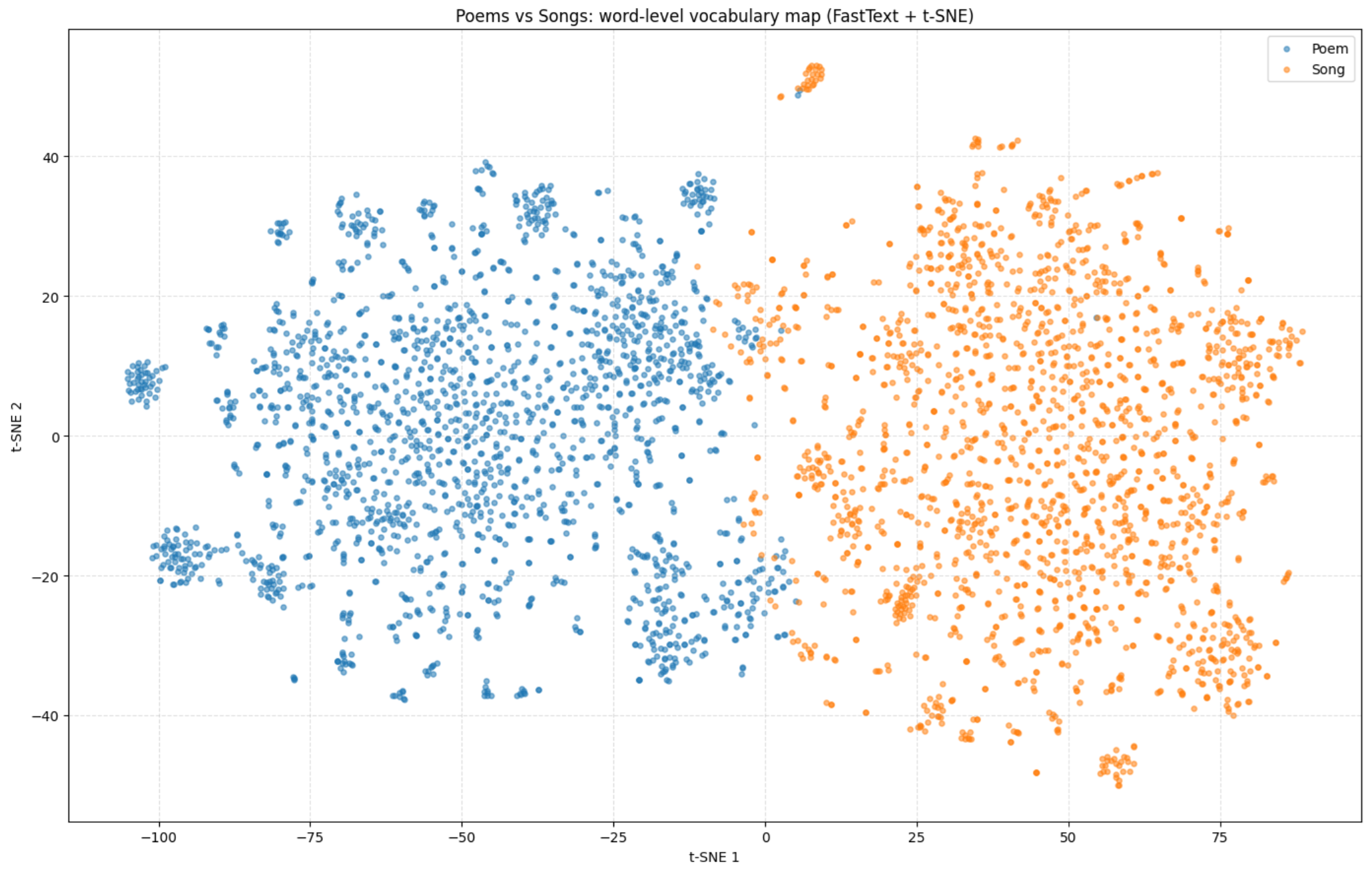}
\caption{Word-level vocabulary map contrasting poems and song lyrics.}
\label{fig:type-tsne}
\end{figure}

Figure~\ref{fig:type-tsne} contrasts poems and song lyrics at the word level. The visualisation shows a clear separation between poetic and lyrical vocabularies once shared high-frequency items are removed. Poetic vocabulary forms a compact and internally cohesive region, consistent with conventionalised lexical choices associated with literary poetry. In contrast, song vocabulary occupies a broader and more fragmented region of the space, suggesting greater lexical diversity and the coexistence of multiple expressive strategies shaped by performance, repetition, and colloquial usage.

Taken together, these visualisations point to a layered lexical structure in Tarab: a highly distinct Classical Arabic stratum, a central and connective MSA layer, and regional dialects and song-specific vocabularies that combine shared lexical material with clusters of strongly distinctive items. This structure highlights the potential of Tarab as a resource for studying lexical variation, register interaction, and dialect-aware representation learning in Arabic NLP.

\section{Artist and Poet Coverage}

The Tarab corpus includes 34,239 unique song titles and 54,927 unique poem titles, reflecting the cultural diversity of Arabic musical and poetic heritage across both modern and historical contexts. In contrast to the original Habibi corpus \cite{elhaj2020habibi}, which ranked artists using raw verse frequency alone, Tarab adopts a more balanced ranking approach that accounts for multiple dimensions of contribution. This reduces bias toward prolific artists with short or repetitive works, as well as poets with unusually long or formulaic compositions.

Specifically, we compute a composite contribution score that equally weights three factors: productivity (number of songs or poems), textual volume (total word count), and dataset presence (total number of verses). Together, these measures capture both the breadth and depth of an artist's or poet's contribution to the corpus. The score for each artist or poet is computed as:
\begin{center}
\resizebox{\columnwidth}{!}{$
\text{score} = \frac{1}{3}
\left(
\frac{\text{words}}{\max(\text{words})} +
\frac{\text{verses}}{\max(\text{verses})} +
\frac{\text{works}}{\max(\text{works})}
\right)
$}
\end{center}
where \textit{works} refers to songs for lyric artists and poems for poets. Tables \ref{tab:artists} and \ref{tab:poets} list the most prominent contributors according to this balanced score. The rankings reveal a mixture of modern music figures, such as \textit{Fayruz} and \textit{Muhammad Abdu}, alongside canonical poets from the Abbasid and Ottoman periods, including \textit{al-Sharif al-Radi} and \textit{Abu al-Ala al-Maarri}. This distribution highlights the cultural depth of the Tarab corpus and its suitability for research in diachronic stylistics, authorship studies, and cultural analytics.

\begin{table}[h]
\centering
\small
\resizebox{\columnwidth}{!}{%
\begin{tabular}{lrrr}
\hline
Artist & Songs & Words & Verses \\
\hline
Fayruz & 681 & 124,268 & 21,920 \\
Muhammad Abdu & 567 & 119,520 & 21,094 \\
Talal Maddah & 523 & 93,836 & 13,926 \\
Asala Nasri & 371 & 80,318 & 13,890 \\
Amr Diyab & 402 & 74,660 & 11,726 \\
Abd Allah al-Ruwaishid & 413 & 68,320 & 12,062 \\
Abd al-Majid Abd Allah & 307 & 68,640 & 12,346 \\
Kazim al-Sahir & 343 & 65,002 & 11,656 \\
Rashid al-Majid & 360 & 65,271 & 10,352 \\
Rabih Saqr & 376 & 60,505 & 9,368 \\
\hline
\end{tabular}
}
\caption{Top lyric artists ranked by a balanced contribution score with equal weighting of songs, words, and verses.}
\label{tab:artists}
\end{table}

\begin{table}[h]
\centering
\small
\resizebox{\columnwidth}{!}{%
\begin{tabular}{lrrr}
\hline
Poet & Poems & Words & Verses \\
\hline
Ibn Nubata al-Misri & 1,726 & 129,088 & 26,567 \\
al-Sharif al-Radi & 677 & 147,974 & 31,953 \\
Abu al-Ala al-Maarri & 1,609 & 113,018 & 23,576 \\
Abd al-Ghani al-Nabulusi & 962 & 120,823 & 26,508 \\
al-Sharif al-Murtada & 587 & 130,799 & 27,561 \\
Ahmad Muharram & 437 & 130,615 & 27,399 \\
Abd al-Ghaffar al-Akhras & 378 & 112,462 & 23,669 \\
Ibn al-Saati & 519 & 102,802 & 21,225 \\
Safiyy al-Din al-Hilli & 898 & 86,857 & 18,236 \\
\hline
\end{tabular}
}
\caption{Top poets ranked by a balanced contribution score with equal weighting of poems, words, and verses.}
\label{tab:poets}
\end{table}

\section{Ethical and Legal Considerations}
\label{sec:ethics}

Tarab is intended for research use. The corpus contains text extracted from publicly accessible sources, including an openly released poetry dataset and lyric text from Kaggle\footnote{\url{https://www.kaggle.com/datasets/ahmedabelal/arabic-poetry}}, as well as material from the Habibi corpus \cite{elhaj2020habibi}. No audio, recordings, or musical compositions are included. Because lyrics and some modern poetic texts may be subject to copyright, we distribute Tarab with an explicit research-oriented usage statement and provide a takedown mechanism for rights holders. The release package is designed to support computational analysis of linguistic and stylistic patterns rather than to substitute access to original works. The dataset is publicly available on HuggingFace \url{https://huggingface.co/datasets/drelhaj/Tarab}.

\section{Limitations and Future Work}

While Tarab provides broad coverage of Arabic creative language, it is not without limitations. First, temporal metadata is coarse-grained for parts of the corpus, particularly for heritage poetry, where association with historical eras is used in place of precise dates. This limits fine-grained diachronic analysis at the year or decade level. Second, although Tarab captures substantial dialectal diversity, dialect labels are assigned at the verse or work level and do not account for intra-textual mixing or gradual register shifts within individual songs or poems. Similarly, stylistic categories such as musical style or performance tradition are maintained separately from the core schema and are not exhaustively annotated across the entire dataset. Finally, the corpus focuses on verse-level textual representation and does not encode higher-level musical, prosodic, or performance features that are central to many forms of Arabic song. As a result, Tarab is best suited to linguistic and stylistic analysis rather than full multimodal or musicological study. Future work could address these limitations by enriching temporal metadata where feasible, expanding auxiliary annotations related to style and performance, and developing benchmark tasks that leverage Tarab's coverage of dialect, genre, and historical depth. Future work could also explore controlled extensions of the corpus that support evaluation of downstream NLP tasks such as dialect identification, authorship attribution, and stylistic transfer.

\section{Conclusion}
This paper introduces the Tarab corpus, a large-scale resource of Arabic creative language that brings together song lyrics and poetry across more than fourteen centuries, multiple genres, and a wide range of linguistic varieties, and is publicly available at \url{https://huggingface.co/datasets/drelhaj/Tarab}. By adopting the verse as a unified analytical unit and separating dialect, origin, and stylistic practice in its design, Tarab enables analyses that are difficult to support using existing Arabic corpora. Through detailed coverage statistics and lexical analyses, we showed that Tarab captures substantial dialectal diversity, clear genre differentiation, and a layered lexical structure spanning Classical Arabic, MSA, and regional varieties. The corpus also preserves cultural depth by representing both canonical poets and contemporary artists, providing a balanced view of Arabic creative production across time. Tarab is intended as a reusable resource for research in Arabic NLP, computational sociolinguistics, and digital humanities, supporting tasks such as dialect modelling, authorship analysis, stylistic variation, and representation learning. Future work could extend the corpus with richer temporal metadata, additional stylistic annotations, and task-specific benchmarks, further strengthening its role as a reference resource for Arabic creative language.

\bibliographystyle{lrec2026-natbib}
\bibliography{custom}

@book{habash2010introduction,
  title={Introduction to Arabic natural language processing},
  author={Habash, Nizar Y},
  year={2010},
  publisher={Morgan \& Claypool Publishers}
}

@article{zaidan-callisonburch-2014-arabic,
  title={Arabic dialect identification},
  author={Zaidan, Omar F and Callison-Burch, Chris},
  journal={Computational Linguistics},
  volume={40},
  number={1},
  pages={171--202},
  year={2014},
  publisher={MIT Press One Rogers Street, Cambridge, MA 02142-1209, USA journals-info~…}
}

@inproceedings{el2018arabic,
  title={Arabic dialect identification in the context of bivalency and code-switching},
  author={El-Haj, Mahmoud and Rayson, Paul and Aboelezz, Mariam},
  booktitle={Proceedings of the 11th International Conference on Language Resources and Evaluation, Miyazaki, Japan.},
  pages={3622--3627},
  year={2018},
  organization={European Language Resources Association}
}

@inproceedings{bouamor-etal-2019-madar,
  title={The MADAR Arabic dialect corpus and lexicon},
  author={Bouamor, Houda and Habash, Nizar and Salameh, Mohammad and Zaghouani, Wajdi and Rambow, Owen and Abdulrahim, Dana and Obeid, Ossama and Khalifa, Salam and Eryani, Fadhl and Erdmann, Alexander and others},
  booktitle={Proceedings of the eleventh international conference on language resources and evaluation (LREC 2018)},
  year={2018}
}

@article{mubarak-etal-2021-arabic,
  title={Arabic dialect identification in the wild},
  author={Abdelali, Ahmed and Mubarak, Hamdy and Samih, Younes and Hassan, Sabit and Darwish, Kareem},
  journal={arXiv preprint arXiv:2005.06557},
  year={2020}
}

@inproceedings{mohammad-etal-2016-arabic,
  title={A compact Arabic lexical semantics language resource based on the theory of semantic fields},
  author={Attia, Mohamed and Rashwan, Mohsen and Ragheb, Ahmed and Al-Badrashiny, Mohamed and Al-Basoumy, Husein and Abdou, Sherif},
  booktitle={International Conference on Natural Language Processing},
  pages={65--76},
  year={2008},
  organization={Springer}
}

@article{alhazmi2024code,
  title={Code-mixing unveiled: Enhancing the hate speech detection in Arabic dialect tweets using machine learning models},
  author={Alhazmi, Ali and Mahmud, Rohana and Idris, Norisma and Mohamed Abo, Mohamed Elhag and Eke, Christopher Ifeanyi},
  journal={Plos one},
  volume={19},
  number={7},
  pages={e0305657},
  year={2024},
  publisher={Public Library of Science San Francisco, CA USA}
}

@inproceedings{obeid2020camel,
  title={CAMeL tools: An open source python toolkit for Arabic natural language processing},
  author={Obeid, Ossama and Zalmout, Nasser and Khalifa, Salam and Taji, Dima and Oudah, Mai and Alhafni, Bashar and Inoue, Go and Eryani, Fadhl and Erdmann, Alexander and Habash, Nizar},
  booktitle={Proceedings of the twelfth language resources and evaluation conference},
  pages={7022--7032},
  year={2020}
}

@article{darwish-2014-arabizi,
  title={Arabizi detection and conversion to Arabic},
  author={Darwish, Kareem},
  journal={arXiv preprint arXiv:1306.6755},
  year={2013}
}

@inproceedings{elhaj2020habibi,
  title={Habibi-a multi dialect multi national Arabic song lyrics corpus},
  author={El-Haj, Mahmoud},
  booktitle={Proceedings of the Twelfth Language Resources and Evaluation Conference},
  pages={1318--1326},
  year={2020}
}

@book{habash2014multidialectal,
  title={A multidialectal parallel corpus of Arabic},
  author={Habash, Nizar and Bouamor, Houda and Oflazer, Kemal},
  year={2014},
  publisher={Carnegie Mellon University}
}

@inproceedings{el2024multilingual,
  title={The Multilingual Corpus of World’s Constitutions (MCWC)},
  author={El-Haj, Mo and Ezzini, Saad},
  booktitle={Proceedings of the 6th Workshop on Open-Source Arabic Corpora and Processing Tools (OSACT) with Shared Tasks on Arabic LLMs Hallucination and Dialect to MSA Machine Translation@ LREC-COLING 2024},
  pages={57--66},
  year={2024}
}

@article{bojanowski2017enriching,
  title={Enriching word vectors with subword information},
  author={Bojanowski, Piotr and Grave, Edouard and Joulin, Armand and Mikolov, Tomas},
  journal={Transactions of the association for computational linguistics},
  volume={5},
  pages={135--146},
  year={2017},
  publisher={MIT Press One Rogers Street, Cambridge, MA 02142-1209, USA journals-info~…}
}

@article{al2020meter,
  title={Meter classification of Arabic poems using deep bidirectional recurrent neural networks},
  author={Al-Shaibani, Maged S and Alyafeai, Zaid and Ahmad, Irfan},
  journal={Pattern Recognition Letters},
  volume={136},
  pages={1--7},
  year={2020},
  publisher={Elsevier}
}

@article{qarah2024arapoembert,
  title={AraPoemBERT: A pretrained language model for Arabic poetry analysis},
  author={Qarah, Faisal},
  journal={arXiv preprint arXiv:2403.12392},
  year={2024}
}

@article{antoun-etal-2020-arabert,
  title={Arabert: Transformer-based model for arabic language understanding},
  author={Antoun, Wissam and Baly, Fady and Hajj, Hazem},
  journal={arXiv preprint arXiv:2003.00104},
  year={2020}
}

@article{el2013kalimat,
  title={KALIMAT a multipurpose Arabic Corpus},
  author={El-Haj, Mahmoud and Koulali, Rim},
  journal={Culture},
  volume={2},
  pages={1--359},
  year={2013}
}

@article{mutawa2025determining,
  title={Determining the meter of classical Arabic poetry using deep learning: a performance analysis},
  author={Mutawa, AM and Alrumaih, Ayshah},
  journal={Frontiers in Artificial Intelligence},
  volume={8},
  pages={1523336},
  year={2025},
  publisher={Frontiers Media SA}
}

@misc{parker2011arabic,
  author       = {Parker, Robert and Graff, David and Chen, Ke and Kong, Junbo and Maeda, Kazuaki},
  title        = {Arabic Gigaword Fifth Edition},
  howpublished = {Linguistic Data Consortium, Catalog Number LDC2011T11},
  year         = {2011},
  url          = {https://catalog.ldc.upenn.edu/LDC2011T11},
  note         = {ISBN 1-58563-595-2}
}

@inproceedings{zerrouki-bouamor-2019-osian,
  title={OSIAN: Open source international Arabic news corpus-preparation and integration into the CLARIN-infrastructure},
  author={Zeroual, Imad and Goldhahn, Dirk and Eckart, Thomas and Lakhouaja, Abdelhak},
  booktitle={Proceedings of the fourth arabic natural language processing workshop},
  pages={175--182},
  year={2019}
}

@inproceedings{zaidan-callisonburch-2011-arabic,
  title={The arabic online commentary dataset: an annotated dataset of informal arabic with high dialectal content},
  author={Zaidan, Omar and Callison-Burch, Chris},
  booktitle={Proceedings of the 49th Annual Meeting of the Association for Computational Linguistics: Human Language Technologies},
  pages={37--41},
  year={2011}
}

@book{openiti2019,
  title={Open Islamicate Texts Initiative (OpenITI), 2016 [Rese{\~n}a]},
  author={Padillo-Saoud, Abdenour},
  year={2019},
  publisher={Universidad Nacional de Educaci{\'o}n a Distancia (Espa{\~n}a)}
}

@article{otakar2014,
  title={Shamela: A large-scale historical Arabic corpus},
  author={Belinkov, Yonatan and Magidow, Alexander and Romanov, Maxim and Shmidman, Avi and Koppel, Moshe},
  journal={arXiv preprint arXiv:1612.08989},
  year={2016}
}

@inproceedings{boudlal2010alkhalil,
  title={Alkhalil morpho sys1: A morphosyntactic analysis system for arabic texts},
  author={Boudlal, Abderrahim and Lakhouaja, Abdelhak and Mazroui, Azzeddine and Meziane, Abdelouafi and Bebah, MOAO and Shoul, Mostafa},
  booktitle={International Arab conference on information technology},
  pages={1--6},
  year={2010},
  organization={Elsevier Science Inc New York, NY}
}

@article{boudchiche2017alkhalil,
  title={AlKhalil Morpho Sys 2: A robust Arabic morpho-syntactic analyzer},
  author={Boudchiche, Mohamed and Mazroui, Azzeddine and Bebah, Mohamed Ould Abdallahi Ould and Lakhouaja, Abdelhak and Boudlal, Abderrahim},
  journal={Journal of King Saud University-Computer and Information Sciences},
  volume={29},
  number={2},
  pages={141--146},
  year={2017},
  publisher={Elsevier}
}
\end{document}